\documentclass[10pt,twocolumn,letterpaper]{article}

\usepackage{cvpr}
\usepackage{times}
\usepackage{color}
\usepackage{graphicx}
\usepackage{amsmath}
\usepackage{amsthm}
\usepackage{algorithm}
\usepackage{algorithmic}
\usepackage{subfigure}
\usepackage{url}
\usepackage{bm}
\usepackage{bbm}
\usepackage{multirow}
 \usepackage{url}
\theoremstyle{plain}

\usepackage[pagebackref=true,breaklinks=true,letterpaper=true,colorlinks,bookmarks=false]{hyperref}

\cvprfinalcopy 

\def\cvprPaperID{1478} 

\ifcvprfinal\fi
\begin{document}


\title{$\mathbf{D^3}$: Deep Dual-Domain Based Fast Restoration of JPEG-Compressed Images}

\author{Zhangyang Wang\dag, Ding Liu\dag, Shiyu Chang\dag, Qing Ling\ddag, Yingzhen Yang\dag, and Thomas S. Huang\dag\thanks{Zhangyang Wang and Thomas Huang's research works are
supported in part by US Army Research Office under grant number W911NF-15-1-0317. Qing Ling's research is supported by CAS grant XDA06011203.}
 \\
\dag Beckman Institute, University of Illinois at Urbana-Champaign, Urbana, IL 61801, USA\\
\ddag Department of Automation, University of Science and Technology of China, Hefei, 230027, China
\\
{\tt\small \{zwang119, dingliu2, chang87, yyang58, t-huang1\}@illinois.edu} $\qquad$ {\tt\small qingling@mail.ustc.edu.cn}
}
\maketitle

\begin{abstract}

In this paper, we design a Deep Dual-Domain ($\mathbf{D^3}$) based
fast restoration model to remove artifacts of JPEG compressed
images. It leverages the large learning capacity of deep networks,
as well as the problem-specific expertise that was hardly
incorporated in the past design of deep architectures. For the
latter, we take into consideration both the prior knowledge of the
JPEG compression scheme, and the successful practice of the
sparsity-based dual-domain approach. We further design the One-Step Sparse Inference (1-SI) module, as an efficient and light-weighted feed-forward approximation of sparse coding. Extensive experiments
verify the superiority of the proposed $D^3$ model over several
state-of-the-art methods. Specifically, our best model is capable
of outperforming the latest deep model for around 1 dB in PSNR,
and is 30 times faster.

\end{abstract}

\section{Introduction}

In visual communication and computing systems, the most common
cause of image degradation is arguably compression. Lossy
compression, such as JPEG \cite{JPEG} and HEVC-MSP \cite{HEVC}, is
widely adopted in image and video codecs for saving both bandwidth
and in-device storage. It exploits inexact approximations for
representing the encoded content compactly. Inevitably, it will
introduce undesired complex artifacts, such as blockiness, ringing
effects, and blurs. They are usually caused by the discontinuities
arising from batch-wise processing, the loss of high-frequency
components by coarse quantization, and so on. These artifacts not
only degrade perceptual visual quality, but also adversely affect
various low-level image processing routines that take compressed
images as input \cite{Dong}.

As practical image compression methods are not information
theoretically optimal \cite{Liu}, the resulting compression code
streams still possess residual redundancies, which makes the
restoration of the original signals possible. Different from
general image restoration problems, compression artifact
restoration has problem-specific properties that can be utilized
as powerful priors. For example, JPEG compression first
divides an image into 8 $\times$ 8 pixel blocks, followed by
discrete cosine transformation (DCT) on every block. Quantization
is applied on the DCT coefficients of every block, with pre-known
quantization levels \cite{JPEG}. Moreover, the
compression noises are more difficult to model than other common
noise types. In contrast to the tradition of assuming noise to be
white and signal independent \cite{KSVD}, the non-linearity of
quantization operations makes quantization noises non-stationary
and signal-dependent.

Various approaches have been proposed to suppress compression
artifacts. Early works \cite{Liu5,Liu16} utilized filtering-based
methods to remove simple artifacts. Data-driven methods were then considered to avoid inaccurate
empirical modeling of compression degradations. Sparsity-based
image restoration approaches have been discussed in
\cite{chang2014reducing, choi2013learning, Dong12,
 Xianming, rothe2015efficient}
to produce sharpened images, but they are often accompanied with
artifacts along edges, and unnatural smooth regions. In \cite{Liu}, Liu et.al. proposed a sparse coding process carried out jointly in the DCT and pixel
domains, to simultaneously exploit residual redundancies of JPEG
code streams and sparsity properties of latent images. More recently, Dong et. al. \cite{Dong} first introduced deep learning
techniques \cite{imagenet} into this problem, by
specifically adapting their SR-CNN model in \cite{SRCNN}. However,
it does not incorporate much problem-specific prior knowledge.


The time constraint is often stringent in image
or video codec post-processing scenarios. Low-complexity or even
real-time attenuation of compression artifacts is highly desirable
\cite{shen1999real}. The inference process of traditional
approaches, for example, sparse coding, usually involves iterative
optimization algorithms, whose inherently sequential structure as
well as the data-dependent complexity and latency often constitute
a major bottleneck in the computational efficiency \cite{LISTA}.
Deep networks benefit from the feed-forward structure and enjoy
much faster inference. However, to maintain their competitive performances, deep networks show demands for increased width (numbers of
filters) and depth (number of layers), as well as smaller strides,
all leading to growing computational costs
\cite{he2015convolutional}.


\begin{figure*}[htbp]
\centering
\begin{minipage}{0.97\textwidth}
\centering{
\includegraphics[width=\textwidth]{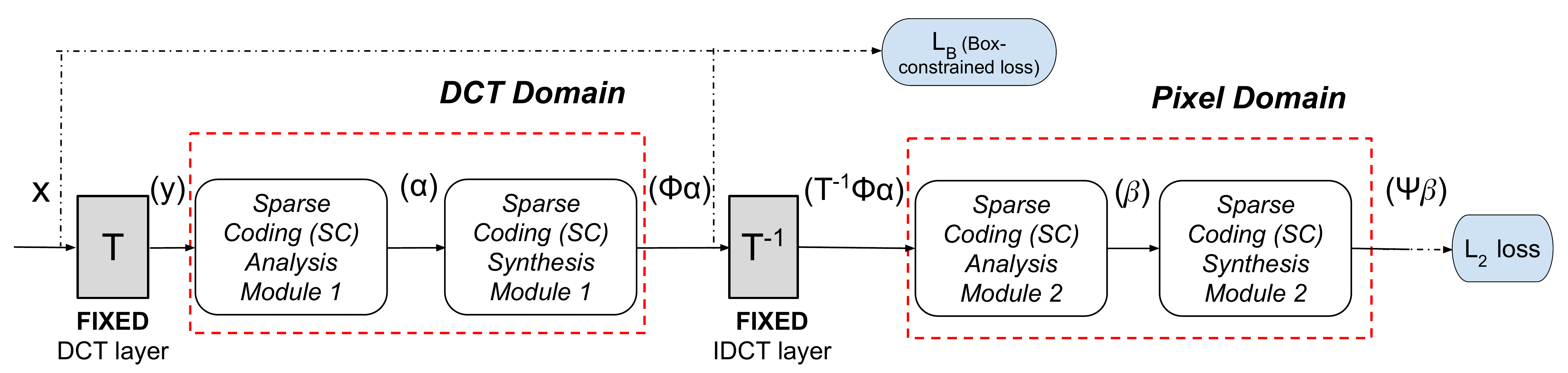}
}\end{minipage}
\caption{The illustration of Deep Dual-Domain ($\mathbf{D^3}$) based model (all subscripts are omitted for simplicity). The black solid lines denote the network inter-layer connections, while the black dash lines connect to the loss functions. The two red dash-line boxes depict the two stages that incorporate DCT and pixel domain sparsity priors, respectively. The two grey blocks denote constant DCT and IDCT layers, respectively. The notations within parentheses along the pipeline are to remind the corresponding variables in (\ref{dual}).}
\label{DDD}
\end{figure*}

In the paper, we focus on removing artifacts in JPEG compressed
images. Our major innovation is to explicitly combine both
\textbf{the prior knowledge in the JPEG compression scheme} and
\textbf{the successful practice of dual-domain sparse coding}
\cite{Liu}, for designing a task-specific deep architecture. Furthermore, we introduce a One-Step Sparse Inference
\textbf{(1-SI)} module, that acts as a highly efficient and light-weighted approximation of the sparse coding inference \cite{Reckless}. 1-SI
also reveals important inner connections between sparse coding and
deep learning. The proposed model, named Deep Dual-Domain
($\mathbf{D^3}$) based fast restoration, proves to be more
effective and interpretable than general deep models. It gains
remarkable margins over several state-of-the-art methods, in terms
of both \textbf{restoration performance} and \textbf{time
efficiency}.

\section{Related Work}

Our work is inspired by the prior wisdom in \cite{Liu}. Most
previous works restored compressed images in either the pixel
domain \cite{KSVD} or the DCT domain \cite{JPEG} solely. However,
an isolated quantization error of one single DCT coefficient is propagated to all pixels of the same block. An aggressively quantized DCT coefficient can further produce structured errors in
the pixel-domain that correlate to the latent signal. On the other
hand, the compression process sets most high frequency
coefficients to zero, making it impossible to recover details from
only the DCT domain. In view of their complementary characteristics, the dual-domain model
was proposed in \cite{Liu}. While the spatial redundancies in the
pixel domain were exploited by a learned dictionary \cite{KSVD},
the residual redundancies in the DCT domain were also utilized to
directly restore DCT coefficients. In this way, quantization noises
were suppressed without propagating errors. The final objective
(see Section 3.1) is a combination of DCT- and pixel-domain sparse
representations, which could cross validate each other.

To date, deep learning \cite{imagenet} has shown impressive
results on both high-level and low-level vision problems \cite{Zhaowen, wang2015self}. The
SR-CNN proposed by Dong et al. \cite{SRCNN} showed the great
potential of end-to-end trained networks in image super resolution (SR).
Their recent work \cite{Dong} proposed a four-layer convolutional
network that was tuned based on SR-CNN, named Artifacts Reduction
Convolutional Neural Networks (AR-CNN), which was
effective in dealing with various compression artifacts.



In \cite{LISTA}, the authors leveraged fast trainable regressors
and constructed feed-forward network approximations of the learned
sparse models. By turning sparse coding into deep networks, one
may expect faster inference, larger learning capacity, and better
scalability. Similar views were adopted in
\cite{PAMI2015} to develop a fixed-complexity algorithm for solving structured sparse and
robust low rank models. The paper \cite{unfold} summarized the
methodology of ``deep unfolding''. \cite{Zhaowen}
proposed deeply improved sparse coding for SR, which can
be incarnated as an end-to-end neural network. Lately, \cite{AAAI16} proposed Deep $\ell_0$ Encoders, to model $\ell_0$ sparse approximation as feed-forward neural networks. \cite{SDM} further extended the same ``task-specific'' strategy to graph-regularized $\ell_1$ approximation. Our task-specific
architecture shares similar spirits with these works.



\section{Deep Dual-Domain ($\mathbf{D^3}$) based Restoration}

\subsection{Sparsity-based Dual-Domain Formulation}

We first review the sparsity-based dual-domain restoration model
established in \cite{Liu}. Considering a training set of \textbf{uncompressed} images, pixel-domain blocks \{$\hat{x}_i$\} $\in R^m$ (vectorized from a $\sqrt{m} \times \sqrt{m}$ patch; $m$ = 64 for JPEG) are drawn for training, along with their \textbf{quantized} DCT coefficient blocks
\{$y_i$\}$\in R^m$. For each (JPEG-coded) input $x_t$ $\in R^m$, two dictionaries
$\bm{\Phi}\in R^{m \times p_\Phi}$ and $\bm{\Psi} \in R^{m \times p_\Psi}$ ($p_\Phi$ and $p_\Psi$ denote the dictionary sizes) are constructed from training data
\{$y_i$\} and \{$\hat{x}_i$\}, in the DCT and pixel domains,
respectively, via locally adaptive feature selection and
projection. The following optimization model is then solved during the testing stage:
\begin{equation}
\begin{array}{l}\label{dual}
\min_{\{\alpha, \beta\}} ||y_t - \bm{\Phi} \alpha||_2^2 + \lambda_1 ||\alpha||_1 \\   \qquad  \quad  +  \lambda_2 ||T^{-1}  \bm{\Phi} \alpha - \bm{\Psi} \beta||_2^2 + \lambda_3 ||\beta||_1,\\
 \qquad \quad s.t.  \quad q^L \preceq \bm{\Phi} \alpha \preceq
 q^U.
\end{array}
\end{equation}
where $y_t \in R^m$ is the DCT coefficient block for $x_t$. $\alpha \in R^{p_\Phi}$
and $\beta \in R^{p_\Psi}$ are sparse codes in the DCT and pixel domains,
respectively. $T^{-1}$ denotes the inverse discrete cosine
transform (IDCT) operator.  $\lambda_1$, $\lambda_2$ and
$\lambda_3$ are positive scalars. One noteworthy point is the
inequality constraint, where $q^L$ and $q^U$ represents the
(pre-known) quantization intervals according to the JPEG
quantization table \cite{JPEG}. The constraint incorporates the
important side information and further confines the solution
space. Finally, $\bm{\Psi} \beta$ provides an estimate of the original uncompressed pixel block $\hat{x}_t$.

Such a sparsity-based dual-domain model (\ref{dual}) exploits
residual redundancies (e,g, inter-DCT-block correlations) in the
DCT domain without spreading errors into the pixel domain, and at
the same time recovers high-frequency information driven by a
large training set. However, note that the inference process of
(\ref{dual}) relies on iterative algorithms, and is computational
expensive. Also in (\ref{dual}), the three parameters $\lambda_1$,
$\lambda_2$ and $\lambda_3$ have to be manually tuned. The authors
of \cite{Liu} simply set them all equal, which may hamper the performance. In
addition, the dictionaries $\bm{\Phi}$ and $\bm{\Psi}$ have to be
individually learned for each patch, which allows for extra
flexibility but also brings in heavy computation load.

\subsection{$\mathbf{D^3}$: A Feed-Forward Network Formulation}

%

In training, we have the pixel-domain blocks \{$x_i$\} after JPEG
compression, as well as the original blocks \{$\hat{x}_i$\}. During testing, for an input compressed block $x_t$, our goal is to estimate the original $\hat{x}_t$, using the redundancies in both DCT and pixel domains, as well as
JPEG prior knowledge.


As illustrated in Fig. \ref{DDD}, the input $x_t$ is first
transformed into its DCT coefficient block $y_t$, by feeding through the
constant 2-D DCT matrix layer $T$. The subsequent two layers aim
to enforce DCT domain sparsity, where we refer to the concepts of
analysis and synthesis dictionaries in sparse coding \cite{Lei}.
The Sparse Coding (SC) Analysis Module 1 is implemented to solve
the following type of sparse inference problem in the DCT domain
($\lambda$ is a positive coefficient):
\begin{equation}
\begin{array}{l}\label{alpha}
\min_\alpha \frac{1}{2}||y_t - \bm{\Phi} \alpha||_2^2 + \lambda ||\alpha||_1.
\end{array}
\end{equation}
The Sparse Coding (SC) Synthesis Module 1 outputs the
DCT-domain sparsity-based reconstruction in (\ref{dual}), i.e.,
$\bm{\Phi} \alpha$.

The intermediate output $\bm{\Phi} \alpha$ is further constrained by
an auxiliary loss, which encodes the inequality constraint in
(\ref{dual}): $ q^L \preceq \bm{\Phi} \alpha \preceq q^U$. We
design the following \textbf{signal-dependent, box-constrained
\cite{box} loss}:
\begin{equation}
\begin{array}{l}\label{box}
L_B(\bm{\Phi} \alpha, x) = ||[\bm{\Phi} \alpha - q^U(x)]_+||_2^2 + ||[q^L(x) - \bm{\Phi} \alpha]_+||_2^2.
\end{array}
\end{equation}
Note it takes not only $\bm{\Phi} \alpha$, but also $x$ as inputs, since the actual JPEG quantization interval [$q^L$, $q^U$] depends on $x$. The operator $[\quad]_+$ keeps the nonnegative elements unchanged while setting others to zero. Eqn. (\ref{box}) will thus only penalize the coefficients falling out of the quantization interval.

After the constant IDCT matrix layer $T^{-1}$, the
DCT-domain reconstruction $\bm{\Phi} \alpha$ is transformed back
to the pixel domain for one more sparse representation. The SC
Analysis Module 2 solves ($\gamma$ is a positive coefficient):
\begin{equation}
\begin{array}{l}\label{beta}
\min_\beta \frac{1}{2}||T^{-1}  \bm{\Phi} \alpha - \bm{\Psi} \beta||_2^2 + \gamma ||\beta||_1,
\end{array}
\end{equation}
while the SC Synthesis Module 2 produces the final pixel-domain
reconstruction $\bm{\Psi} \beta$. Finally, the $L_2$ loss between $\bm{\Psi}
\beta$ and $\hat{x}_i$ is enforced.

Note that in the above, we try to correspond the intermediate
outputs of $\mathbf{D^3}$ with the variables in
(\ref{dual}), in order to help understand the close analytical
relationship between the proposed deep architecture with the sparse
coding-based model. That does not necessarily imply any exact
numerical equivalence, since $\mathbf{D^3}$ allows for end-to-end
learning of all parameters (including $\lambda$ in (\ref{alpha})
and $\gamma$ in (\ref{beta})). However, we will see in experiments
that such enforcement of the specific problem structure improves
the network performance and efficiency remarkably. In addition,
the above relationships remind us that the deep model could be well initialized from the sparse coding components.

\begin{figure*}[tbp]
\centering
\begin{minipage}{0.225\textwidth}
\centering \subfigure[Original] {
\includegraphics[width=\textwidth]{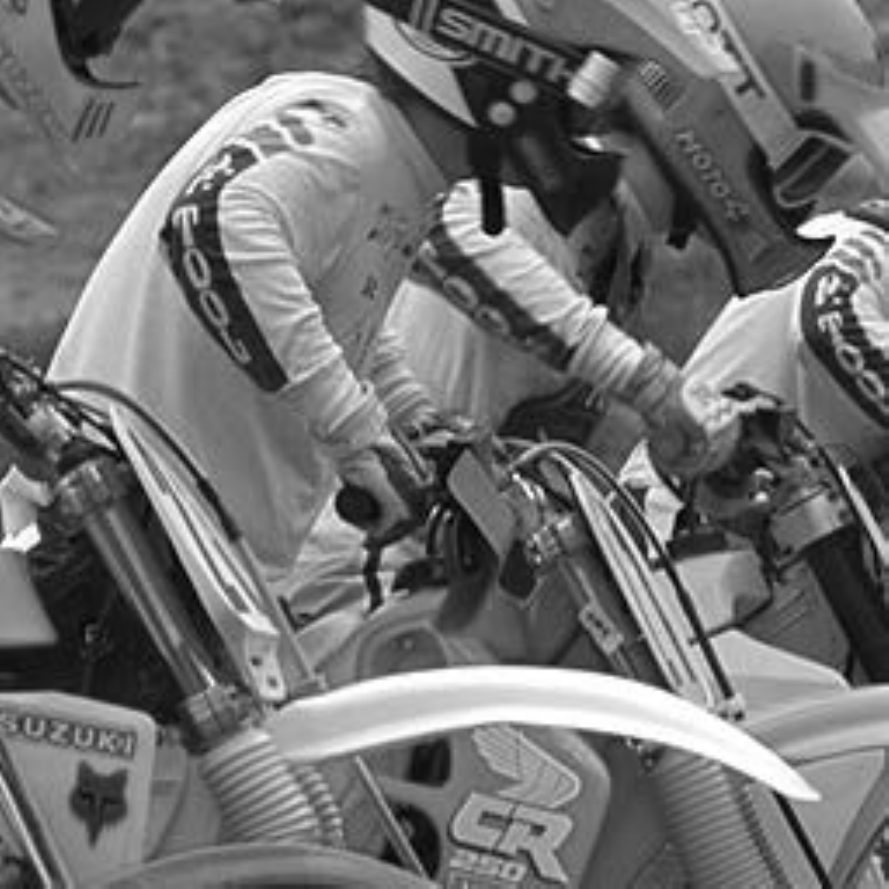}
}\end{minipage}
\begin{minipage}{0.225\textwidth}
\centering \subfigure[Compressed (PSNR = 21.72 dB)] {
\includegraphics[width=\textwidth]{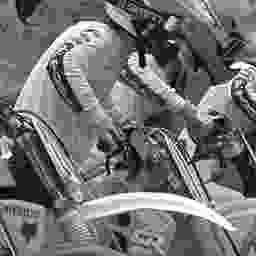}
}\\
\centering \subfigure[S-D$^2$ (PSNR = 22.87 dB)] {
\includegraphics[width=\textwidth]{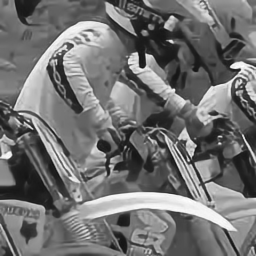}
}\end{minipage}
\begin{minipage}{0.225\textwidth}
\centering \subfigure[AR-CNN (PSNR = 23.27 dB)] {
\includegraphics[width=\textwidth]{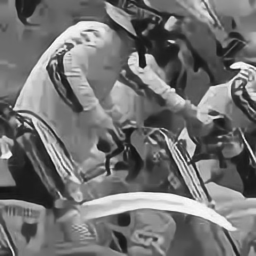}
}\\
\centering \subfigure[D$^3$-128 (PSNR = 23.94 dB)] {
\includegraphics[width=\textwidth]{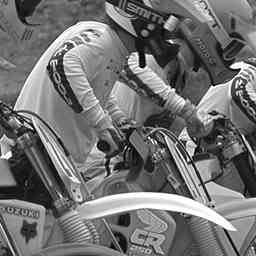}
}\end{minipage}
\begin{minipage}{0.225\textwidth}
\centering \subfigure[D$^3$-256 (PSNR = 24.30 dB)] {
\includegraphics[width=\textwidth]{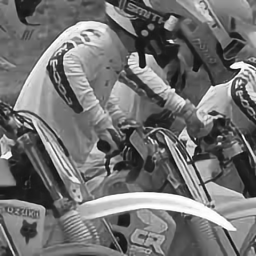}
}\\
\centering \subfigure[D-Base-256 (PSNR = 23.48 dB)] {
\includegraphics[width=\textwidth]{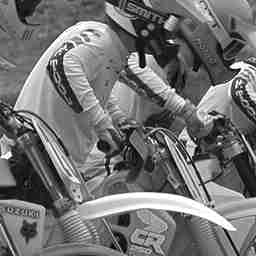}
}\end{minipage}
\caption{Visual comparison of various methods on \textit{Bike} at Q = 5. The corresponding PSNR values (in dB) are also shown.}
\label{bike}
\end{figure*}

\subsection{One-Step Sparse Inference Module}


The implementation of SC Analysis and Synthesis Modules appears to be the core of $\mathbf{D^3}$. While the synthesis process is naturally feed-forward by multiplying the dictionary, it is less straightforward to transform the sparse analysis (or inference) process into a feed-forward network. 

We take (\ref{alpha}) as an example, while the same solution applies
to (\ref{beta}). Such a sparse inference problem could be solved
by the iterative shrinkage and thresholding algorithm (ISTA)
\cite{iterative}, each iteration of which updates as follows:
\begin{equation}
\begin{array}{l}\label{ISTA}
\bm{\alpha}^{k+1} = s_{\lambda}(\bm{\alpha}^k + \bm{\Phi}^T (y_t - \bm{\Phi} \bm{\alpha}^k)),
\end{array}
\end{equation}
where $\bm{\alpha}^k$ denotes the intermediate result of the $k$-th iteration, and where $s_{\lambda}$ is an element-wise shrinkage function ($\mathbf{u}$ is a vector and $\mathbf{u}_i$ is its $i$-th element, $i = 1, 2, ..., p$):
\begin{equation}
\begin{array}{l}\label{threshold}
[s_{\lambda}(\mathbf{u})]_i = \text{sign}(\mathbf{u}_i)[|\mathbf{u}_i| - \lambda_i]_{+}.
\end{array}
\end{equation}
The learned ISTA (LISTA) \cite{LISTA} parameterized encoder
further proposed a natural network implementation of ISTA. The
authors time-unfolded and truncated (\ref{ISTA}) into a fixed
number of stages (more than 2), and then jointly tuned all
parameters with training data, for a good feed-forward
approximation of sparse inference. The similar unfolding
methodology has been lately exploited in \cite{unfold},
\cite{PAMI2015}, \cite{sprechmann2013supervised}.

In our work, we launch a more aggressive approximation, by only
keeping one iteration of (\ref{ISTA}), leading to a One-Step
Sparse Inference (\textbf{1-SI}) Module. Our major motivation lies
in the same observation as in \cite{Dong} that overly deep
networks could adversely affect the performance in low-level
vision tasks. Note that we have two SC Analysis modules where the
original LISTA applies, and two more SC Synthesis modules (each
with one learnable layer). Even only two iterations are kept as in
\cite{LISTA}, we end up with a six-layer network, that suffers
from both difficulties in training \cite{Dong} and fragility in
generalization \cite{srivastava2014dropout} for this task.

A 1-SI module takes the following simplest form:
\begin{equation}
\begin{array}{l}\label{1step}
\bm{\alpha}= s_{\lambda}(\bm{\Phi} y_t),
\end{array}
\end{equation}
which could be viewed as first passing through a fully-connected
layer ($\bm{\Phi}$), followed by neurons that take the form of
$s_{\lambda}$. We further rewrite (\ref{threshold}) as
\cite{Zhaowen} did\footnote{In (\ref{threshold1}), we slightly
abuse notations, and set $\lambda$ to be a vector of the same
dimension as $\mathbf{u}$, in order for extra element-wise
flexibility.}:
\begin{equation}
\begin{array}{l}\label{threshold1}
[s_{\lambda}(\mathbf{u})]_i = \lambda_i \cdot \text{sign}(\mathbf{u}_i)(|\mathbf{u}_i|/\lambda_i - 1)_{+} = \lambda_i s_1(\mathbf{u}_i/\lambda_i)
\end{array}
\end{equation}
Eqn. (\ref{threshold1}) indicates that the original neuron with
trainable thresholds can be decomposed into two linear scaling
layers plus a unit-threshold neuron. The weights of the two
scaling layers are diagonal matrices defined by $\bm{\theta}$ and
its element-wise reciprocal, respectively. The unit-threshold
neuron $s_1$ could in essence be viewed as a double-sided and
translated variant of ReLU \cite{imagenet}.

\begin{figure}[htbp]
\centering
\begin{minipage}{0.45\textwidth}
\centering{
\includegraphics[width=\textwidth]{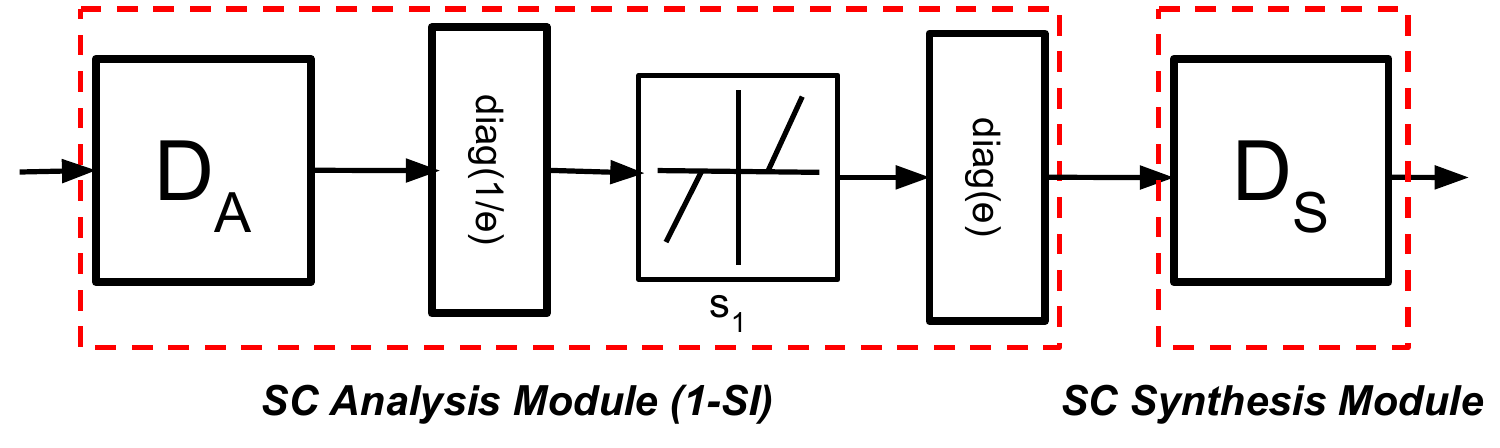}
}\end{minipage}
\caption{The illustration of SC Analysis and Synthesis Modules. The former is implemented by the proposed 1-SI module (\ref{1step}). Both $D_A$ and $D_S$ are fully-connected layers, while diag($\theta$) and diag($1/\theta$) denotes the two diagonal scaling layers.}
\label{1SI}
\end{figure}

A related form to (\ref{1step}) was obtained in \cite{Reckless} on
a different case of non-negative sparse coding. The authors
studied its connections with the soft-threshold feature for
classification, but did not correlate it with network architectures.


\begin{figure*}[tbp]
\centering
\begin{minipage}{0.225\textwidth}
\centering \subfigure[Original] {
\includegraphics[width=\textwidth]{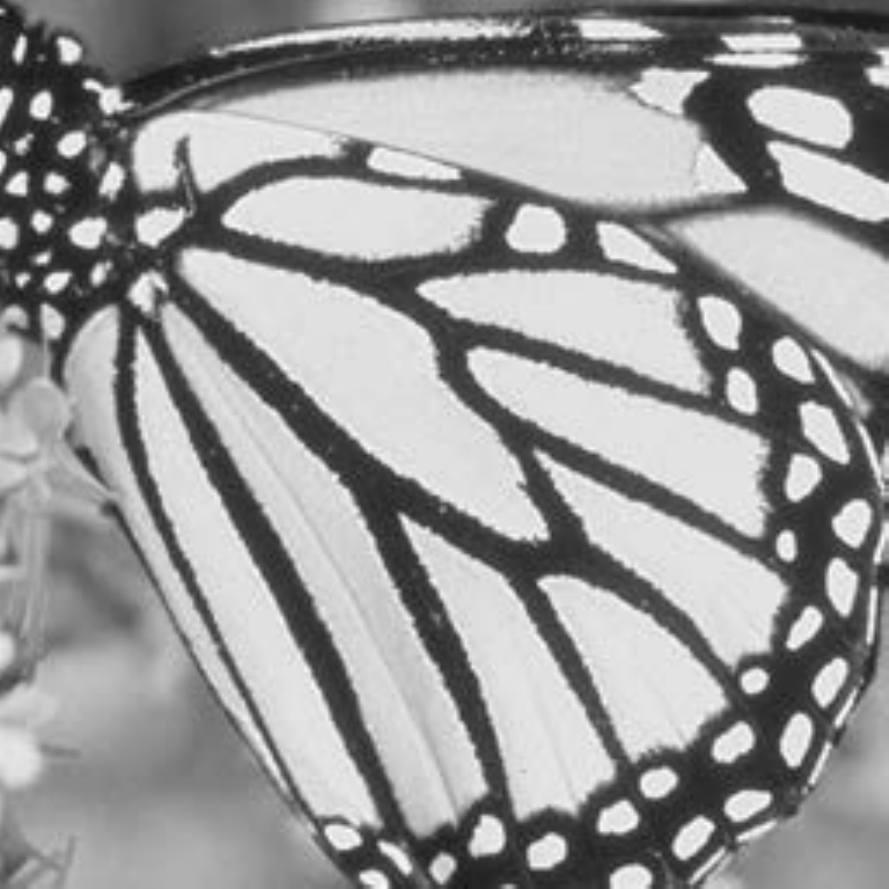}
}\end{minipage}
\begin{minipage}{0.225\textwidth}
\centering \subfigure[Compressed (PSNR = 22.65 dB)] {
\includegraphics[width=\textwidth]{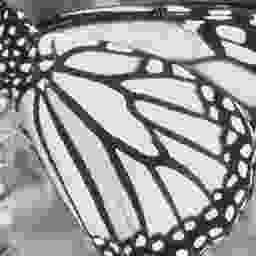}
}\\
\centering \subfigure[S-D$^2$ (PSNR = 24.87 dB)] {
\includegraphics[width=\textwidth]{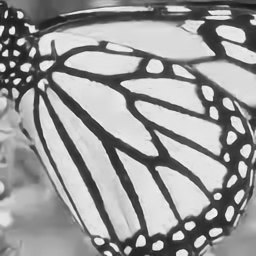}
}\end{minipage}
\begin{minipage}{0.225\textwidth}
\centering \subfigure[AR-CNN (PSNR = 25.81 dB)] {
\includegraphics[width=\textwidth]{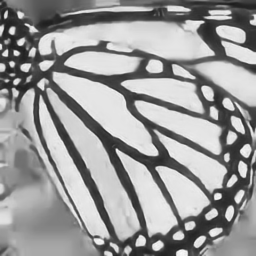}
}\\
\centering \subfigure[D$^3$-128 (PSNR = 24.74 dB)] {
\includegraphics[width=\textwidth]{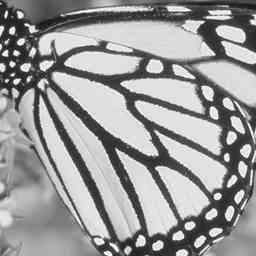}
}\end{minipage}
\begin{minipage}{0.225\textwidth}
\centering \subfigure[D$^3$-256 (PSNR = 26.30 dB)] {
\includegraphics[width=\textwidth]{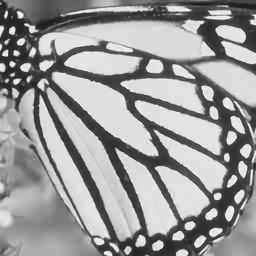}
}\\
\centering \subfigure[D-Base-256 (PSNR = 24.28 dB)] {
\includegraphics[width=\textwidth]{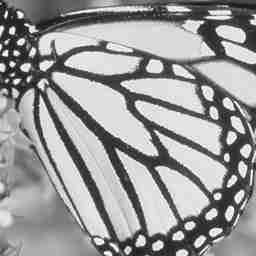}
}\end{minipage}
\caption{Visual comparison of various methods on \textit{Monarch} at Q = 5. The corresponding PSNR values (in dB) are also shown.}
\label{butterfly}
\end{figure*}

\subsection{Model Overview}

By plugging in the 1-SI module (\ref{1step}), we are ready to
obtain the SC Analysis and Synthesis Modules, as in Fig.
\ref{1SI}. By comparing Fig. \ref{1SI} with Eqn. (\ref{alpha}) (or
(\ref{beta})), it is easy to notice the analytical relationships
between $D_A$ and $\bm{\Phi}^T$ (or $\bm{\Psi}^T$), $D_S$ and
$\bm{\Phi}$ (or $\bm{\Psi}$), as well as $\theta$ and $\lambda$
(or $\gamma$). In fact, those network hyperparamters could be well
initialized from the sparse coding parameters, which
could be obtained easily. The entire $D^3$ model, consisting
of four learnable fully-connected weight layers (except for the
diagonal layers), are then trained from end to end \footnote{From
the analytical perspective, $\mathbf{D}_S$ is the transpose of
$\mathbf{D}_A$, but we untie them during training for larger
learning capability.}.

In Fig. \ref{1SI}, we intentionally do not combine $\theta$ into
$\mathbf{D}_A$ layer (also $1/\theta$ into $\mathbf{D}_S$ layer ),
for the reason that we still wish to keep $\theta$ and $1/\theta$
layers tied as element-wise reciprocal. That proves to have
positive implications in our experiments. If we absorb the two
diagonal layers into $\mathbf{D}_A$ and $\mathbf{D}_S$, Fig.
\ref{1SI} is reduced to two fully connected weight matrices,
concatenated by one layer of hidden neurons (\ref{threshold1}).
However, keeping the ``decomposed'' model architecture facilitates
the incorporation of problem-specific structures.




\subsection{Complexity Analysis}

From the clear correspondences between the sparsity-based
formulation and the $D^3$ model, we immediately derive the dimensions of weight layers, as in Table
\ref{dim}.

\begin{figure*}[tbp]
\centering
\begin{minipage}{0.225\textwidth}
\centering \subfigure[Original] {
\includegraphics[width=\textwidth]{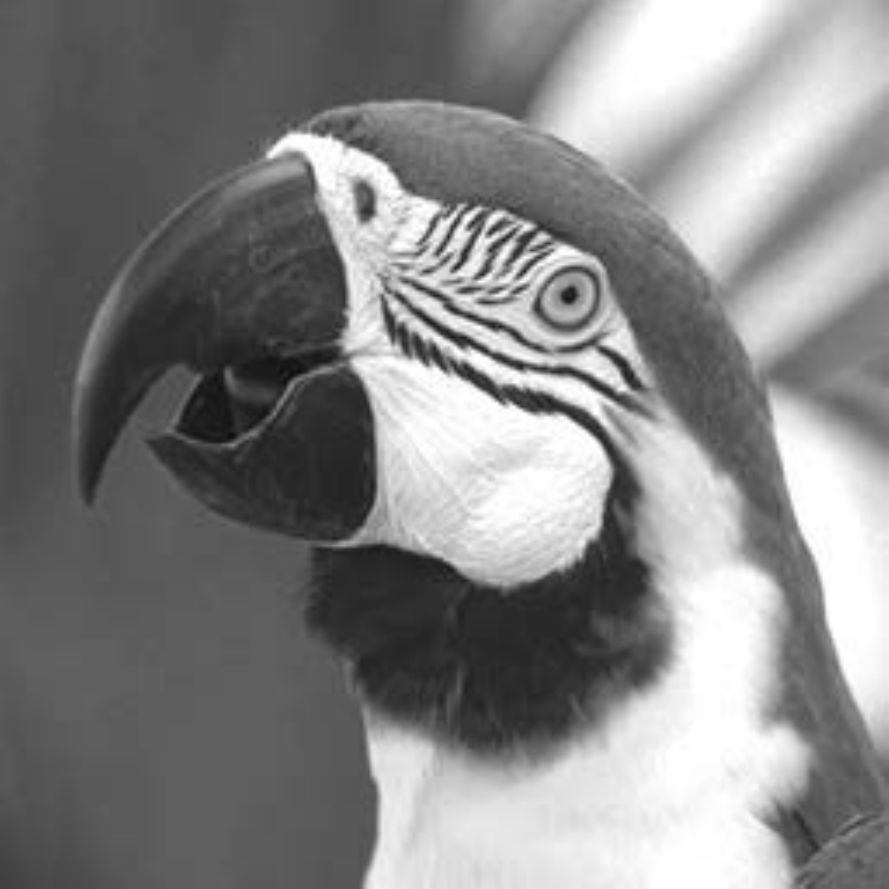}
}\end{minipage}
\begin{minipage}{0.225\textwidth}
\centering \subfigure[Compressed (PSNR = 26.15 dB)] {
\includegraphics[width=\textwidth]{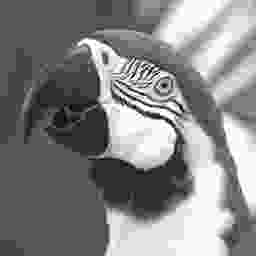}
}\\
\centering \subfigure[S-D$^2$ (PSNR = 27.92 dB)] {
\includegraphics[width=\textwidth]{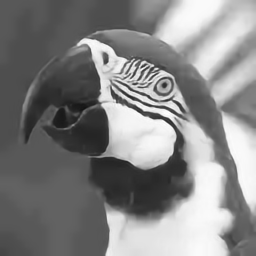}
}\end{minipage}
\begin{minipage}{0.225\textwidth}
\centering \subfigure[AR-CNN (PSNR = 28.20 dB)] {
\includegraphics[width=\textwidth]{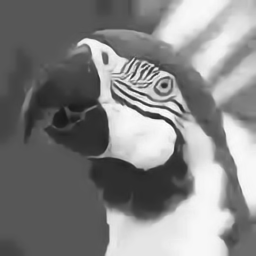}
}\\
\centering \subfigure[D$^3$-128 (PSNR = 27.52 dB)] {
\includegraphics[width=\textwidth]{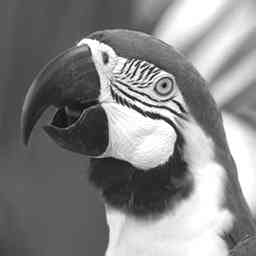}
}\end{minipage}
\begin{minipage}{0.225\textwidth}
\centering \subfigure[D$^3$-256 (PSNR = 28.84 dB)] {
\includegraphics[width=\textwidth]{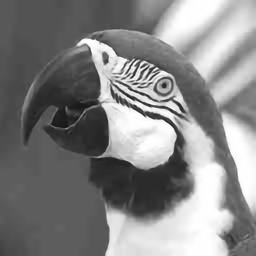}
}\\
\centering \subfigure[D-Base-256 (PSNR = 27.21 dB)] {
\includegraphics[width=\textwidth]{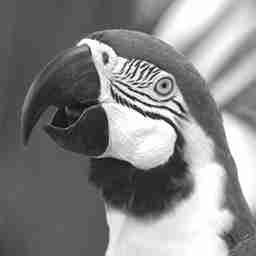}
}\end{minipage}
\caption{Visual comparison of various methods on \textit{Parrots} at Q = 5. The corresponding PSNR values are also shown.}
\label{parrot}
\end{figure*}

 \begin{table}[htbp]
  \small
  \begin{center}
 \caption{Dimensions of all layers in the $D^3$ model}
 \label{dim}
 \begin{tabular}{|c|c|c|c|}
 \hline
Layer & $\mathbf{D}_A$ &  $\mathbf{D}_S$ &  diag($\theta$) \\
\hline
Stage I (DCT Domain) & $p_\Phi \times m$ & $m \times p_\Phi$ & $p_\Phi$ \\
\hline
Stage II (Pixel Domain) & $p_\Psi \times m$ & $m \times p_\Psi$ & $p_\Psi$ \\
\hline
 \end{tabular}
 \end{center}
 \end{table}
 \vspace{-1em}

\subsubsection{Time Complexity}

During training, deep learning with the aid of gradient descent
scales linearly in time and space with the number of training
samples. We are primarily concerned with the time complexity
during testing (inference), which is more relevant to practical
usages. Since all learnable layers in the $D^3$ model are
fully-connected, the inference process of $D^3$ is nothing more
than a series of matrix multiplications. The multiplication times
are counted as: $p_\Phi m$ ($D_A$ in Stage I) + $2p_\Phi$ (two
diagonal layers) + $p_\Phi m$ ($D_S$ in Stage I)  + $p_\Psi m$
($D_A$ in Stage II) + $2p_\Psi$ (two diagonal layers) + $p_\Psi m$
($D_S$ in Stage II). The 2D DCT and IDCT each takes $\frac{1}{2}
m\log(m)$ multiplications \cite{JPEG} . Therefore, the total
inference time complexity of $D^3$ is:
\begin{equation}
\begin{array}{l}\label{D3complexity}
C_{D^3} = 2 (p_\Phi+ p_\Psi) (m + 1) + m\log(m) \approx 2m(p_\Phi+ p_\Psi).
\end{array}
\end{equation}
The complexity could also be expressed as $O(p_\Phi+ p_\Psi)$.

It is obvious that the sparse coding inference \cite{Liu} has
dramatically higher time complexity. We are also interested in the
inference time complexity of other competitive deep models,
especially AR-CNN \cite{Dong}. For their fully convolutional
architecture, the total complexity \cite{he2015convolutional} is:
\begin{equation}
\begin{array}{l}\label{convcomplexity}
C_{conv} = \sum_{l=1}^d n_{l-1} \cdot s_l^2 \cdot n_l \cdot m_l^2,
\end{array}
\end{equation}
where $l$ is the layer index, $d$ is the total depth, $n_l$ is the
number of filters in the $l$-th layer, $s_l$ is the spatial size
of the filter, and $m_l$ is the spatial size of the output feature
map.

The theoretical time complexities in (\ref{D3complexity}) and
(\ref{convcomplexity}) do not represent  the actual running time,
as they depend on different configurations and can be sensitive to
implementations and hardware. Yet, our actual running time scales
nicely with those theoretical results.

 \subsubsection{Parameter Complexity}
The total number of free parameters in $D^3$ is:
\begin{equation}
\begin{array}{l}\label{D3para}
N_{D^3} = 2 p_\Phi m + p_\Phi + 2 p_\Psi m + p_\Psi = 2 (p_\Phi+
p_\Psi) (m + 1).
\end{array}
\end{equation}
As a comparison, the AR-CNN model \cite{Dong} contains:
\begin{equation}
\begin{array}{l}\label{convpara}
N_{conv} =  \sum_{l=1}^d n_{l-1} \cdot n_l \cdot s_l^2.
\end{array}
\end{equation}

 \begin{table*}[htbp]
 \small
 \begin{center}
 \caption{The average results of PSNR (dB), SSIM, PSNR-B (dB) on the LIVE1 dataset.}
 \label{allcompare}
 \begin{tabular}{|c|c|c|c|c|c|c|c|}
 \hline
& & Compressed & S-D$^2$ &  AR-CNN & D$^3$-128 & D$^3$-256& D-Base-256 \\
  \hline
$\multirow{3}{*}{Q = 5}$ & PSNR & 24.61 & 25.83 & 26.64  & 26.26 & \textbf{27.37} & 25.83 \\
\cline{2-8}
 & SSIM & 0.7020 & 0.7170& 0.7274 & 0.7203 & \textbf{0.7303} & 0.7186  \\
 \cline{2-8}
& PSNR-B & 22.01 & 25.64 & 26.46   & 25.86 & \textbf{26.95} & 25.51  \\
  \hline
$\multirow{3}{*}{Q = 10}$ & PSNR & 27.77 & 28.88 & 29.03   & 28.62 & \textbf{29.96} & 28.24 \\
\cline{2-8}
 & SSIM & 0.7905 & 0.8195& 0.8218  & 0.8198 & \textbf{0.8233} & 0.8161  \\
 \cline{2-8}
& PSNR-B & 25.33 & 27.96& 28.76 & 28.33 & \textbf{29.45} &  27.57 \\
  \hline
$\multirow{3}{*}{Q = 20}$ & PSNR & 30.07 & 31.62 & 31.30  & 31.20 & \textbf{32.21} & 31.27 \\
\cline{2-8}
 & SSIM & 0.8683 & 0.8830 & 0.8871  & 0.8829 & \textbf{0.8903} & 0.8868 \\
 \cline{2-8}
& PSNR-B & 27.57 & 29.73 & 30.80  & 30.56 & \textbf{31.35} & 29.25 \\
 \hline
  \hline
\#Param &  & $\backslash$ & NA & 106,448 & 33, 280 & 66, 560 & 66, 560 \\
  \hline
 \end{tabular}
 \end{center}
 \end{table*}

\section{Experiments}

\subsection{Implementation and Setting}

We use the disjoint training set (200 images) and test set
(200 images) of BSDS500 database \cite{BSD}, as our training set; its validation set (100 images) is used for validation, which follows \cite{Dong}.
For training the D$^3$ model, we first divide each original image
into overlapped $8 \times 8$ patches, and subtract the pixel
values by 128 as in the JPEG mean shifting process. We then
perform JPEG encoding on them by MATLAB JPEG encoder with a
specific quality factor $Q$, to generate the corresponding
compressed samples. Whereas JPEG works on non-overlapping patches,
we emphasize that the training patches are overlapped and
extracted from arbitrary positions. For a testing image, we sample
$8 \times 8$ blocks with a stride of 4, and apply the D$^3$ model in a
patch-wise manner. For a patch that misaligns with the original
JPEG block boundaries, we find its most similar coding block from
its $16 \times 16$ local neighborhood, whose quantization
intervals are then applied to the misaligned patch. We find this
practice effective and important for removing blocking artifacts
and ensuring the neighborhood consistency. The final result is
obtained via aggregating all patches, with the overlapping
regions averaged.

The proposed networks are implemented using the cuda-convnet
package \cite{imagenet}. We apply a constant learning rate of
0.01, a batch size of 128, with no momentum. Experiments run on a workstation with 12 Intel Xeon
2.67GHz CPUs and 1 GTX680 GPU. The two losses, $L_B$ and $L_2$, are equally weighted. For the parameters in Table
\ref{dim}, $m$ is fixed as 64. We try different values of $p_\Phi$ and $p_\Psi$ in experiments.

Based on the solved Eqn. (\ref{dual}), one could initialize $D_A$, $D_S$, and $\theta$  from $\Phi$, $\Phi^T$ and $\lambda$ in the DCT domain block of Fig. \ref{DDD}, and from $\Psi$, $\Psi^T$ and $\gamma$ in the pixel domain block, respectively. In practice, we find that such an initialization strategy benefits the performances, and usually leads to faster convergence.

We test the quality factor $Q$ = 5, 10, and
20. For each $Q$, we train a dedicated model. We further find the
easy-hard transfer suggested by \cite{Dong} useful. As images of
low $Q$ values (heavily compressed) contain more complex
artifacts, it is helpful to use the features learned from images
of high $Q$ values (lightly compressed) as a starting point. In
practice, we first train the D$^3$ model on JPEG compressed images
with $Q = 20$ (the highest quality). We then initialize the $Q =
10$ model with the $Q = 20$ model, and similarly, initialize $Q =
5$ model from the $Q = 10$ one.

\subsection{Restoration Performance Comparison}
We include the following two relevant, state-of-the-art methods for comparison:
\begin{itemize}
\item \textbf{Sparsity-based Dual-Domain Method (S-D$^2$)}
\cite{Liu} could be viewed as the ``shallow'' counterpart of
D$^3$. It has outperformed most traditional methods
\cite{Liu}, such as BM3D \cite{BM3D} and DicTV
\cite{chang2014reducing}, with which we thus do not compare again.
The algorithm has a few parameters to be manually
tuned. Especially, their dictionary atoms are adaptively selected by a nearest-neighbour type algorithm; the number of selected atoms varies for every testing patch. Therefore, the parameter complexity of
S-D$^2$ cannot be exactly computed.
\item \textbf{AR-CNN} has been
the latest deep model resolving the JPEG compression artifact
removal problem. In \cite{Dong}, the authors show its advantage
over SA-DCT \cite{foi2007pointwise}, RTF \cite{jancsary2012loss},
and SR-CNN \cite{SRCNN}. We adopt the default network
configuration in \cite{Dong}: $s_1$ = 9, $s_2$ = 7, $s_3$ = 1,
$s_4$ = 5; $n_1$ = 64, $n_2$ = 32, $n_3$ = 16, $n_4$ = 1. The
authors adopted the easy-hard transfer in training.
\end{itemize}
For D$^3$, we test $p_\Phi$ =
$p_\Psi$ = 128 and 256 \footnote{from the common experiences of choosing dictionary
sizes \cite{KSVD}}. The resulting D$^3$ models
are denoted as D$^3$-128 and D$^3$-256, respectively.
In addition, to verify the superiority of our task-specific
design, we construct a fully-connected Deep Baseline Model
(D-Base), of the same complexity with D$^3$-256, named D-Base-256.
It consists of four weight matrices of the same dimensions as
D$^3$-256's four trainable layers\footnote{D-Base-256 is a four-layer neural network, performed on the pixel domain, without DCT/IDCT layers. The diagonal layers contain a very small portion of parameters and are ignored here.}. D-Base-256 utilizes ReLU \cite{imagenet} neurons and the dropout technique.

%

We use the 29 images in the LIVE1 dataset \cite{live} (converted to the gray scale) to evaluate both the quantitative and qualitative performances. Three quality assessment criteria: PSNR, structural similarity (SSIM) \cite{SSIM}, and PSNR-B \cite{PSNRB}, are evaluated, the last of which is designed specifically to assess blocky images. The averaged results on the LIVE1 dataset are list in Table \ref{allcompare}. 

Compared to S-D$^2$, both D$^3$-128 and D$^3$-256 gain remarkable advantages, thanks to the end-to-end training as deep architectures. As $p_\Phi$ and $p_\Psi$ grow from 128 to 256, one observes clear improvements in PSNR/SSIM/PSNR-B. D$^3$-256 has outperformed the state-of-the-art ARCNN, for around 1 dB in PSNR. Moreover, D$^3$-256 also demonstrates a notable performance margin over D-Base-256, although they possess the same number of parameters. D$^3$ is thus verified to benefit from its task-specific architecture inspired by the sparse coding process (\ref{dual}), rather than just the large learning capacity of generic deep models. The parameter numbers of different models are compared in the last row of Table \ref{allcompare}. It is impressive to see that D$^3$-256 also takes less parameters than AR-CNN.

We display three groups of visual results, on \textit{Bike}, \textit{Monarch} and \textit{Parrots} images, when $Q$ = 5, in Figs. \ref{bike}, \ref{butterfly} and \ref{parrot}, respectively. AR-CNN tends to generate over-smoothness, such as in the edge regions of butterfly wings and parrot head. S-D$^2$ is capable of restoring sharper edges and textures. The D$^3$ models further reduce the unnatural artifacts occurring in S-D$^2$ results.
Especially, while D$^3$-128 results still suffer from a small amount of visible ringing artifacts, D$^3$-256 not only shows superior in preserving details, but also suppresses artifacts well.


\begin{figure}[h]
\centering
\begin{minipage}{0.15\textwidth}
\centering \subfigure[ PSNR = 22.14 dB] {
\includegraphics[width=\textwidth]{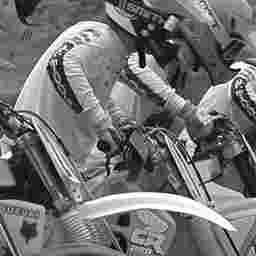}
}\\
\centering \subfigure[ PSNR = 25.14 dB] {
\includegraphics[width=\textwidth]{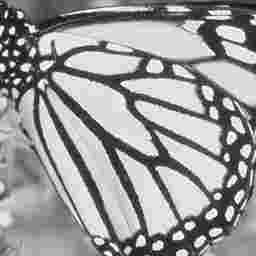}
}\\
\centering \subfigure[PSNR = 26.74 dB] {
\includegraphics[width=\textwidth]{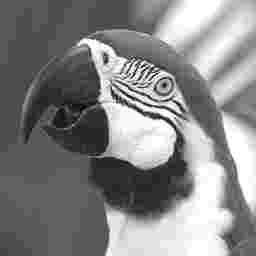}
}\end{minipage}
\begin{minipage}{0.15\textwidth}
\centering \subfigure[ PSNR = 23.42 dB] {
\includegraphics[width=\textwidth]{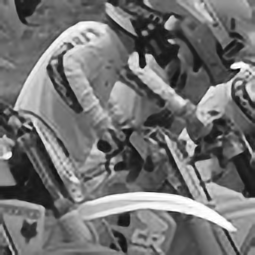}
}\\
\centering \subfigure[ PSNR = 24.85 dB] {
\includegraphics[width=\textwidth]{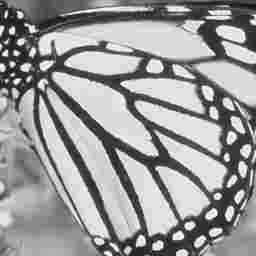}
}\\
\centering \subfigure[PSNR = 27.63 dB] {
\includegraphics[width=\textwidth]{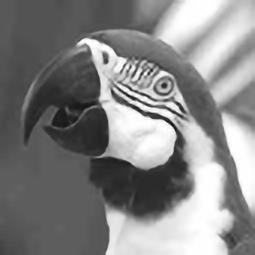}
}\end{minipage}
\begin{minipage}{0.15\textwidth}
\centering \subfigure[ PSNR = 23.80 dB] {
\includegraphics[width=\textwidth]{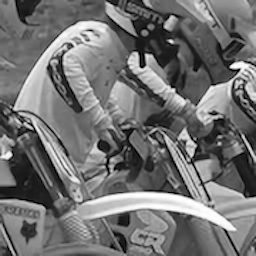}
}\\
\centering \subfigure[ PSNR = 25.63 dB] {
\includegraphics[width=\textwidth]{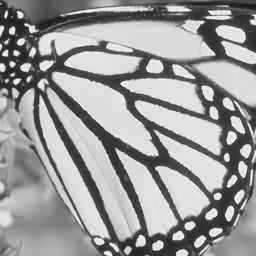}
}\\
\centering \subfigure[PSNR = 28.28 dB] {
\includegraphics[width=\textwidth]{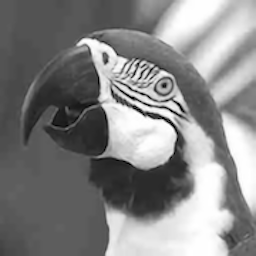}
}\end{minipage}
\caption{Intermediate and comparison results, on \textit{Bike}, \textit{Monarch}, and \textit{Parrot}, at Q = 5: (a) - (c) the intermediate recovery results after the DCT-domain reconstruction; (d) - (f) the results trained with random initialization; (g) - (i) the results trained without the box-constrained loss. PSNR values are reported.}
\label{all}
\end{figure}

\subsection{Analyzing the Impressive Results of D$^3$}

We attribute our impressive recovery of clear fine details, to the combination of our specific pipeline, the  initialization, and the box-constrained loss.


\noindent \textbf{Task-specific and interpretable pipeline}
The benefits of our specifically designed architecture were demonstrated by the comparison experiments to baseline encoders. Further, we provide intermediate outputs of the IDCT layer, i.e., the recovery after the DCT-domain reconstruction. We hope that it helps understand how each component, i.e., the DCT-domain reconstruction or the pixel-domain reconstruction, contributes to the final results. As shown in Fig. \ref{all} (a)-(c), such intermediate reconstruction results contain both sharpened details (see the characters in (a), which become more recognizable), and unexpected noisy patterns (see (a) (b) (c) for the blockiness, and ringing-type noise along edges and textures). It implies that Stage I DCT-domain reconstruction has enhanced the high-frequency features, yet introducing artifacts simultaneously due to quantization noises. Afterwards, Stage II pixel-domain reconstruction performs extra noise suppression and global reconstruction, which leads to the artifact-free and more visually pleasing final results.

\noindent \textbf{Sparse coding-based initialization}
We conjecture that the reason why $D^3$ is more capable in restoring the text on \textit{Bike} and other subtle textures hinges on our sparse coding-based initialization, as an important training detail in $D^3$. To verify that, we re-train $D^3$ with random initialization, with the testing results in Fig. \ref{all} (d)-(f), which turn out to be visually smoother (closer to AR-CNN results). For example, the characters in (d) are now hardly recognizable. We notice that the S-$D^2$ results, as in original Fig. \ref{bike}-\ref{parrot} (c), also presented sharper and more recognizable texts and details than AR-CNN. These observations validate our conjecture. So the next question is, \textbf{why sparse coding helps significantly here}? The quantization process can be considered as as a low-pass filter that cuts off high-frequency information. The dictionary atoms are learned from offline high-quality training images, which contain rich high-frequency information. The sparse linear combination of atoms is thus richer in high-frequency details, which might not necessarily be the case in generic regression (as in deep learning).


\noindent \textbf{Box-constrained loss}
The loss $L_B$ (\ref{box}) acts as another effective regularization. We re-train $D^3$ without the loss, and obtain the results in Fig. \ref{all} (g)-(i). It is observed that the box-constrained loss helps generate details (e.g., comparing characters in (g) with those in Fig. \ref{bike} (f)), by bounding the DCT coefficients, and brings PSNR gains.

\subsection{Running Time Comparison}

The image or video codecs desire highly efficient compression
artifact removal algorithms as the post-processing tool.
Traditional TV and digital cinema business uses frame rate
standards such as 24p (i.e., 24 frames per second), 25p, and 30p.
Emerging standards require much higher rates. For example,
high-end High-Definition (HD) TV systems adopt 50p or 60p; the
Ultra-HD (UHD) TV standard advocates 100p/119.88p/120p; the HEVC
format could reach the maximum frame rate of 300p \cite{url}. To
this end, higher time efficiency is as desirable as improved
performances.

 \begin{table}[htbp]
 \small
 \begin{center}
 \caption{Averaged running time comparison (ms) on LIVE1.}
 \label{timetable}
 \begin{tabular}{|c|c|c|c|c|}
 \hline
& AR-CNN &  D$^3$-128 &  D$^3$-256  & D-Base-256 \\
 \hline
  Q = 5 & 396.76 & 7.62 & 12.20 & 9.85 \\
  \hline
  Q = 10 & 400.34 & 8.84 & 12.79 & 10.27 \\
  \hline
  Q = 20 & 394.61 & 8.42 & 12.02 & 9.97 \\
  \hline
 \end{tabular}
 \end{center}
 \end{table}
\vspace{-1em}



We compare the averaged testing times of AR-CNN and the proposed
D$^3$ models in Table \ref{timetable}, on the LIVE29
dataset, using the same machine and software environment. All running time was collected from GPU tests. Our best model, D$^3$-256, takes approximately 12 ms per image; that is more
than \textbf{30 times faster} than AR-CNN. The speed difference is NOT mainly
caused by the different implementations. Both being completely
feed-forward, AR-CNN relies on the time-consuming convolution
operations while ours takes only a few matrix multiplications.
That is in accordance with the theoretical time complexities
computed from (\ref{D3complexity}) and (\ref{convcomplexity}), too. As
a result, D$^3$-256 is able to process 80p image sequences (or
even higher). To our best knowledge, D$^3$ is the \textbf{fastest}
among all state-of-the-art algorithms, and proves to be a
practical choice for HDTV industrial usage.

\section{Conclusion}
We introduce the D$^3$ model, for the fast restoration of JPEG compressed images. The successful combination of both JPEG prior knowledge and sparse coding expertise has made D$^3$ highly effective and efficient. In the future, we aim to extend the methodology to more related applications.

{\small
\bibliographystyle{ieee}
\bibliography{cvprd4}
}

\end{document}